\documentclass[runningheads]{llncs}
\usepackage[T1]{fontenc}
\usepackage{graphicx}
\usepackage{booktabs}
\usepackage[misc]{ifsym}
\newcommand{\corr}{(\Letter)}

\usepackage{mwe}

\usepackage{nicefrac}
\usepackage{amsmath}
\usepackage{amssymb}
\usepackage{xcolor}
\usepackage{hyperref}
\usepackage{enumitem}
\usepackage{mdframed}
\usepackage{xfakebold}
\usepackage{caption}
\usepackage{wrapfig}
\usepackage{graphicx}
\usepackage{wrapfig}  
\usepackage{booktabs} 
\newcommand{\multiset}[1]{
  \{\!\!\{#1\}\!\!\}
}

\newcommand{\bfA}{\mathbf{A}}
\newcommand{\bfL}{\mathbf{L}}
\newcommand{\bfX}{\mathbf{X}}
\newcommand{\bfD}{\mathbf{D}}
\newcommand{\bfI}{\mathbf{I}}

\newcommand{\fbseries}{\unskip\setBold\aftergroup\unsetBold\aftergroup\ignorespaces}
\newcommand{\setBoldness}[1]{\def\fake@bold{#1}}
\setBoldness{0.5}

\newcommand{\first}[1]{{{\fbseries \textcolor{teal}{#1}}}}
\newcommand{\second}[1]{{{\fbseries \textcolor{violet}{#1}}}}
\newcommand{\third}[1]{{{\fbseries \textcolor{black}{#1}}}}

\begin{document}

\title{Understanding and Improving Laplacian Positional Encodings For Temporal GNNs}

\titlerunning{Understanding Laplacian Positional Encodings For Temporal GNNs}

\author{Yaniv Galron\inst{1} \corr\and Fabrizio Frasca\inst{1} \and Haggai Maron\inst{1,4}\ \and Eran Treister\inst{2} \and Moshe Eliasof \inst{3}}

\authorrunning{Y. Galron et al.}

\institute{
Technion – Israel Institute of Technology \\ \email{yaniv.galron@campus.technion.ac.il}
\and
Ben-Gurion University of the Negev 
\and
University of Cambridge
\and
NVIDIA
}

\maketitle              

\begin{abstract}
Temporal graph learning has applications in recommendation systems, traffic forecasting, and social network analysis. Although multiple architectures have been introduced, progress in positional encoding for temporal graphs remains limited. Extending static Laplacian eigenvector approaches to temporal graphs through the supra-Laplacian has shown promise, but also poses key challenges: high eigendecomposition costs, limited theoretical understanding, and ambiguity about when and how to apply these encodings. 
In this paper, we address these issues by (1) offering a theoretical framework that connects supra-Laplacian encodings to per-time-slice encodings, highlighting the benefits of leveraging additional temporal connectivity, (2) introducing novel methods to reduce the computational overhead, achieving up to 56x faster runtimes while scaling to graphs with 50,000 active nodes, and (3) conducting an extensive experimental study to identify which models, tasks, and datasets benefit most from these encodings. Our findings reveal that while positional encodings can significantly boost performance in certain scenarios, their effectiveness varies across different models.
\end{abstract}

\section{Introduction}

Temporal Graph Neural Networks (TGNNs) have emerged as a state-of-the-art paradigm for learning on dynamic graphs \cite{Rossi2020-we,Kumar2019-cr,Wang2021-tu,Souza2022-uc,Sato2019-uc,Cong2023-nc,trivedi2018representationlearningdynamicgraphs,Huang2023-iq}. By simultaneously capturing evolving temporal dynamics and underlying graph structure, TGNNs have achieved remarkable performance across applications like temporal link prediction \cite{10.1145/3625820,10411773,zhang2024efficientneuralcommonneighbor}, node classification \cite{sun2022dynamicgraphnodeclassification}, and edge classification \cite{ozmen2024recent}.

Positional Encoding (PE) techniques, fundamental to the success of Transformer architectures \cite{Vaswani2017-gl,Heo2024-nz,Dosovitskiy2020-il}, enhance representational capacity by embedding crucial positional information within sequential and temporal data. In static graph contexts, positional encodings---particularly the Laplacian Positional Encoding (LPE) \cite{Belkin2003-cy,Dwivedi2020-ce} derived from the spectral decomposition of the graph Laplacian, have demonstrated significant benefits in injecting structural information and elevating performance across node classification and link prediction tasks \cite{10.5555/3618408.3618777,Dwivedi2020-ce,Lim2022-ia}.

Despite their potential benefits, positional encodings for temporal graphs remain largely underexplored. A recent advancement has been the adaptation of LPEs for TGNNs through the novel application of supra-Laplacian eigenvectors \cite{karmim2024supralaplacian}. The supra-Laplacian \cite{PhysRevE.109.064309,Yang2020-zu,Lin2021-mf,Gomez2013-gg,Sole-Ribalta2013-ha} extends traditional graph Laplacian frameworks by incorporating temporal connectivity between time steps, thereby elegantly capturing both intra-layer structural and inter-layer temporal dynamics. This approach enriches positional encodings with temporal information and has empirically demonstrated improved downstream performance.

However, the adoption of supra-Laplacian based PEs (SLPEs) presents several substantial challenges that warrant a thorough study. First, the theoretical underpinnings and properties of these novel encodings—and their specific relevance to temporal graph learning—remain insufficiently characterized, hampering our understanding of their effectiveness. Second, computing the eigendecomposition of the supra-Laplacian introduces considerable computational overhead due to the increased dimensionality from temporal connections. Third, while initial research \cite{karmim2024supralaplacian} demonstrated promising results with specific transformer-based TGNN architecture and datasets, the generalizability of SLPEs across diverse architectural frameworks and learning tasks remains an open question.

\begin{figure}[t]
    \centering
    \includegraphics[width=1\linewidth]{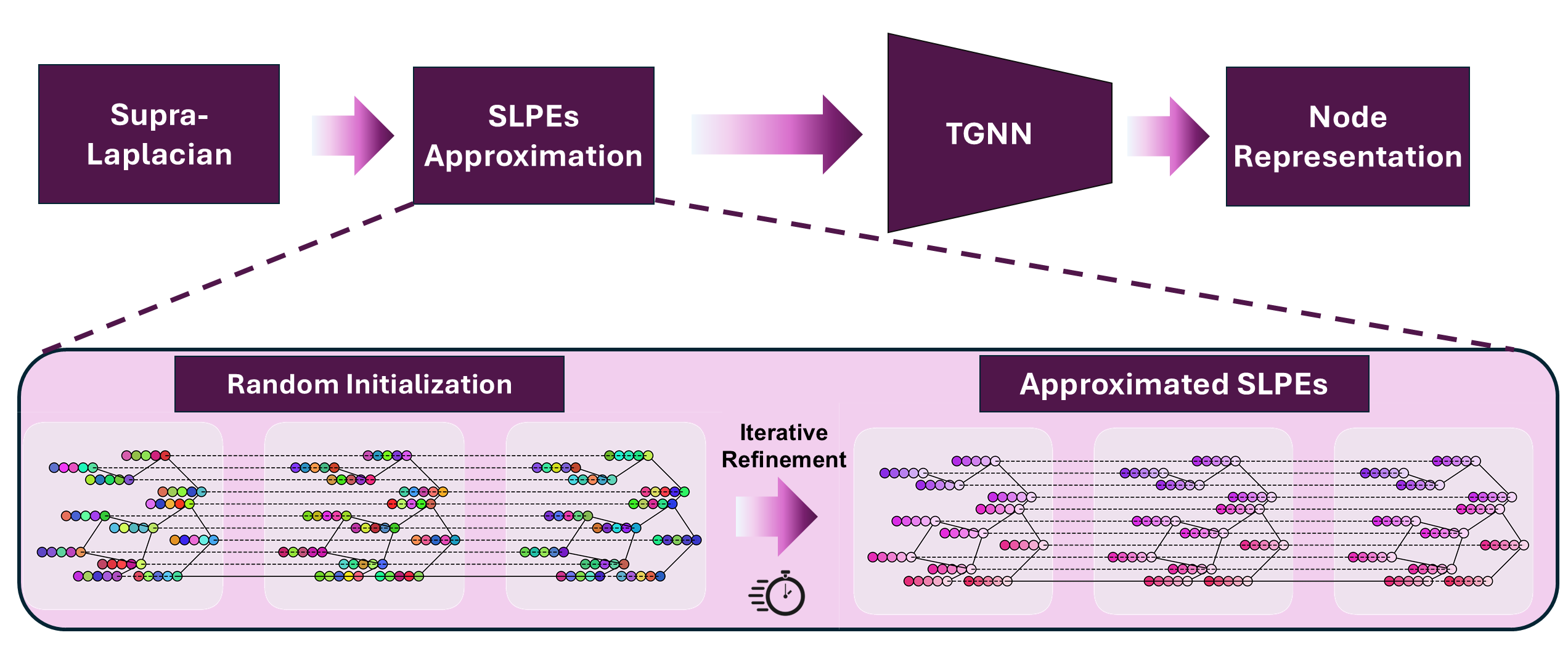}
    \caption{An overview of our proposed fast SLPEs computation procedure. The SLPEs are generated via iterative solvers applied on the supra-Laplacian. Those start with random initialization and apply iterative refinement toward smooth PEs that act as the node representations.}
    \label{fig:overview}
\end{figure}

\textbf{Main Contributions.} This paper systematically addresses the three aforementioned gaps to advance both the theoretical and practical understanding of Laplacian-based PEs for TGNNs:
\begin{enumerate}
\item We develop a theoretical analysis of supra-Laplacian PEs (SLPEs) as compared to single-layer Laplacian PEs (LPEs), and discuss the increased expressive power given by the supra-graph representation.
\item We introduce a computationally efficient framework for calculating SLPEs through approximate eigenvector computation, shown in Figure \ref{fig:overview}.
\item We present extensive empirical evaluations across multiple Laplacian-based PEs, feature initializations, and architectural paradigms (message-passing- and transformer-based TGNNs), culminating in actionable practical guidelines.
\end{enumerate}

\section{Related Work}
We now provide an overview of relevant topics to our work, namely TGNNs and the use of Laplacian Positional Encodings in graph learning.

\textbf{Temporal Graph Neural Networks.} 
TGNNs operate on both Continuous-Time Dynamic Graphs (CTDGs)~\cite{Rossi2020-we,Huang2023-iq,Kumar2019-cr,Souza2022-uc} and Discrete-Time Dynamic Graphs (DTDGs)~\cite{9439502,yang2023dynamicgraphrepresentationlearning}, with efforts to bridge the two domains \cite{Souza2022-uc,Huang2024-dz}.
For DTDGs, early methods like EGCN~\cite{Pareja2019EvolveGCNEG} use a Recurrent Neural Networks approach to apply a Graph Convolutional Network (GCN) over time. HTGN~\cite{Yang2021-rs} leverages hyperbolic geometry to model complex, hierarchical structures in evolving networks. For CTDGs, pioneering methods like DyRep~\cite{trivedi2018representationlearningdynamicgraphs} and JODIE \cite{Kumar2019-cr} process timestamped edge streams, while TGAT~\cite{Xu2020-ns} focuses on inductive representation learning. Temporal Graph Networks (TGNs)~\cite{Rossi2020-we} generalize these approaches, encompassing DyRep, JODIE, and TGAT as special cases. In the context of PEs, \cite{Souza2022-uc} incorporated relative PEs into CTDGs by counting node appearances on temporal walks and~\cite{wang2024dynamicgraphtransformercorrelated} constructed PEs for CTDGs by leveraging the Poisson point process to efficiently estimate personalized interaction intensity.

\textbf{Laplacian-Based Positional Encodings.} Graph Laplacian eigenvectors~\cite{Belkin2003-cy} have gained widespread adoption as effective graph embedding tools. In static graph neural networks, these embeddings encode crucial structural information that demonstrably enhances GNN expressive power~\cite{Dwivedi2020-ce,Rampasek2022-iy,maskey2022generalizedlaplacianpositionalencoding}. The recent work in~\cite{10.5555/3618408.3618777} revealed that approximate eigenvectors—as well as their computation trajectories—can match or surpass the performance of exact eigenvectors. Meanwhile, \cite{Lim2022-ia} developed novel neural architectures invariant to inherent eigenvector symmetries, specifically sign flips and more general basis transformations. Important theoretical challenges were addressed by~\cite{huang2024on}, which investigated the non-uniqueness and instability issues where minor perturbations to the Laplacian can produce substantially different eigenspaces. Building on these advances, \cite{karmim2024supralaplacian} recently extended Laplacian PEs to the temporal graphs for TGNNs, by incorporating the supra-Laplacian into an innovative transformer-based architecture. A theoretical analysis of the supra-Laplacian for applications to graph learning is, however, still missing, as well as their practical effectiveness on other known message-passing-based TGNNs.

\section{Supra-Laplacian PEs in Temporal Graphs} 
In this section, we first introduce the notation used in the paper, and then define the supra-Laplacian and SLPEs that were proposed in the recent work~\cite{karmim2024supralaplacian} to extend Graph Transformers to DTDGs. We consider SLPEs for MPNNs as well.

\textbf{Notations and Definitions.}
\label{sec:notations_definitions}
We follow the setup of~\cite{10.5555/3455716.3455786} where temporal graphs are represented as a sequence of snapshots $\mathbb{G} = \{G_1, ..., G_T\}$. Each snapshot $G_t = (\mathcal{V}_t, \mathcal{E}_t)$ contains nodes $\mathcal{V}_t$ and edges $\mathcal{E}_t$ at time step $t$. Nodes $v_t \in \mathcal{V}_t$ possess feature vectors $\mathbf{h}_{v_t} \in \mathbb{R}^d$, while edges $(u_t,v_t) \in \mathcal{E}_t$ may have associated features $\mathbf{w}_e \in \mathbb{R}^{d_e}$. Collectively, input node features are denoted as $\mathbf{H}_t \in \mathbb{R}^{|\mathcal{V}_t| \times d}$. In addition, we denote by  \(\mathbf{P}_t \in \mathbb{R}^{|\mathcal{V}_t| \times c}\)  the PEs at time $t$. As is standard \cite{Dwivedi2020-ce,10.5555/3618408.3618777}, the PEs are combined with input node features to form an initial representation \(\tilde{\mathbf{H}}_t = [\mathbf{H}_t \| \mathbf{P}_t] \in \mathbb{R}^{|\mathcal{V}_t| \times (d + c)}\), where \(\|\) denotes channel-wise concatenation.

\textbf{Supra-Laplacian and -Adjacency.}
The supra-Laplacian~\cite{PhysRevE.109.064309,Yang2020-zu,Lin2021-mf,Gomez2013-gg,Sole-Ribalta2013-ha} leverages the multi-layer structure of temporal graphs by constructing a block matrix representation of the graph sequence. For a temporal graph \(\mathbb{G}\), the supra-Laplacian matrix \(\mathbf{L}_{\text{supra}} \in \mathbb{R}^{T|\mathcal{V}| \times T|\mathcal{V}|}\) is defined as:
\begin{equation} 
\mathbf{L}_{\text{supra}} = \mathbf{D}_{\text{supra}} - \mathbf{A}_{\text{supra}},
\end{equation}
\noindent where \(\mathbf{A}_{\text{supra}}\) is the supra-adjacency matrix, \(\mathbf{D}_{\text{supra}}\) is the corresponding degree matrix, and $|\mathcal{V}| = \max_{t=1,\ldots,T} |\mathcal{V}_t|$. The supra-adjacency matrix is constructed by placing the adjacency matrices \(\mathbf{A}_t\) of each snapshot \(G_t\) along the diagonal and adding inter-layer edges to model temporal dependencies, defined as:

\begin{equation}
\centering
\mathbf{A}_{\text{supra}} = 
\begin{bmatrix}
\mathbf{A}_1 & \mathbf{B}_{12} & \cdots & \mathbf{B}_{1T} \\
\mathbf{B}_{21} & \mathbf{A}_2 & \cdots & \mathbf{B}_{2T} \\
\vdots & \vdots & \ddots & \vdots \\
\mathbf{B}_{T1} & \mathbf{B}_{T2} & \cdots & \mathbf{A}_T
\end{bmatrix},
\end{equation}
\noindent where \(\mathbf{A}_t \in \mathbb{R}^{|\mathcal{V}| \times |\mathcal{V}|}\) is the adjacency matrix of snapshot \(G_t\), and \(\mathbf{B}_{ij} \in \mathbb{R}^{|\mathcal{V}| \times |\mathcal{V}|}\) represents the inter-layer edges modeling temporal dependencies between snapshots \(G_i\) and \(G_j\). Here, we set \(\mathbf{B}_{ij}\) to be the identity matrix \(\mathbf{I}\) when \(|i - j| = 1\), i.e., the snapshots are connected to their immediate previous and next layers. 

To work with evolving graphs, one can use a subset of the most recent snapshots of $\mathbb{G}$ where the window size represents the number of consecutive time steps or graph snapshots. For simplicity, we consider the window to be of size $T$.

\begin{figure}[t]
    \centering  
    \includegraphics[width=0.8\linewidth]{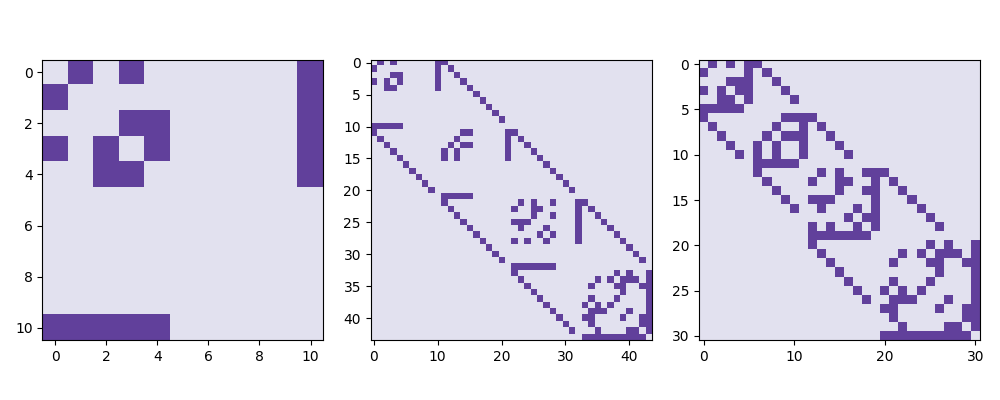}
    \caption{Examples of adjacency matrices. Left: an adjacency matrix of a single snapshot with a global node connected only to the active nodes. Middle: the Supra-Adjacency matrix of shape $|\mathcal{V}|T\times |\mathcal{V}|T$ (middle). Right: the reduced Supra-Adjacency that contains only active nodes per snapshot.}
\label{fig:adj-matrix}
\end{figure}

\textbf{Supra-Laplacian PEs (SLPEs).}\label{paragraph:slpe} These PEs were firstly introduced in \cite{karmim2024supralaplacian} where, for a node \(v\) at time \(t\), they are derived from the eigenvectors of \(\mathbf{L}_{\text{supra}}\) corresponding to the smallest eigenvalues. Let \(\mathbf{X} \in \mathbb{R}^{T|\mathcal{V}| \times k}\) be the matrix of eigenvectors corresponding to the \(k\) smallest eigenvalues of \(\mathbf{L}_{\text{supra}}\). The SLPEs \(\mathbf{P}_t(v)\) for node \(v\) at time \(t\) is given by:

\begin{equation}    
\mathbf{P}_t(v) = \mathbf{X}_{(t-1)|\mathcal{V}| + v, :k},
\end{equation}
where \(\mathbf{X}_{(t-1)|\mathcal{V}| + v, :k}\) extracts the \(k\)-dimensional embedding corresponding to node \(v\) at time \(t\). This encoding captures both the structural and temporal properties of the graph by leveraging the spectral properties of the supra-Laplacian. Furthermore, alongside these eigenvectors, the corresponding smallest eigenvalues of \(\mathbf{L}_{\text{supra}}\) can also be concatenated to add further spectral information about the dynamic graph.

To enhance SLPEs, the work in \cite{karmim2024supralaplacian} proposed two key modifications: (1) Global Node Integration: Each layer is augmented with a global node connected to all active nodes within that layer to better capture layer-wide activity and context (these nodes are considered part of $|\mathcal{V}|$); (2) Isolated Node Removal: Unconnected nodes are removed from the supra-adjacency matrix to eliminate the noise possibly introduced by considering uninformative eigenvectors.

\section{Theoretical Understanding of Supra-Laplacian PEs}

In this section, we enhance the theoretical understanding of SLPEs. First, we show how SLPEs inherently balance intra-layer structure preservation with inter-layer consistency through their smoothness. Second, we analyze the expressiveness advantages of the supra-adjacency matrix over layer-wise methods.

\subsection{Time and Space Smoothness with SLPEs}

We now show that computing the $d$ lowest supra-Laplacian eigenvectors is equivalent to minimizing an objective function that balances the preservation of layer-specific structure while promoting smooth transitions across layers via a penalty term. The proof of the proposition below appears in Appendix \ref{app:supra-smoothness-proof}.

\begin{proposition} \label{prop:smooth} (Supra-Laplacian PEs Smoothness). Let \(\mathbb{G} = \{G_1, G_2, \dots, G_T\}\) be a multilayer graph with \(T\) layers, where each layer \(G_t\) is represented by its adjacency matrix \(\mathbf{A}_t\), the degree matrix \(\mathbf{D}_t\), and the Laplacian matrix \(\mathbf{L}_t\). In addition, \(\mu>0\) is a parameter that controls the weight of inter-layer connections. Then the eigenvectors of the supra-Laplacian matrix associated with $\mathbb{G}$ are the vectors 
$\mathbf{X}^{(t)}\in\mathbb{R}^{|\mathcal{V}|\times k}$ that minimize the following objective function:
\begin{equation}\label{eq:smoothness_obj}
\min_{\mathbf{X}^{(t)}} \sum_{t=1}^T \text{tr}\left({\mathbf{X}^{(t)}}^T \mathbf{L}_t \mathbf{X}^{(t)}\right) + \mu \sum_{t=2}^T\left\| \mathbf{X}^{(t)} - \mathbf{X}^{(t-1)} \right\|_F^2,
\end{equation}
subject to \({\mathbf{X}}^T \mathbf{X} = \mathbf{I}\), where $\bfX$ is the concatenation of all matrices $\mathbf{X}^{(t)}$ and $\|\|_F$ is the Frobenius norm.
\end{proposition}

\begin{figure}[t]
    \centering
    \includegraphics[width=\linewidth]{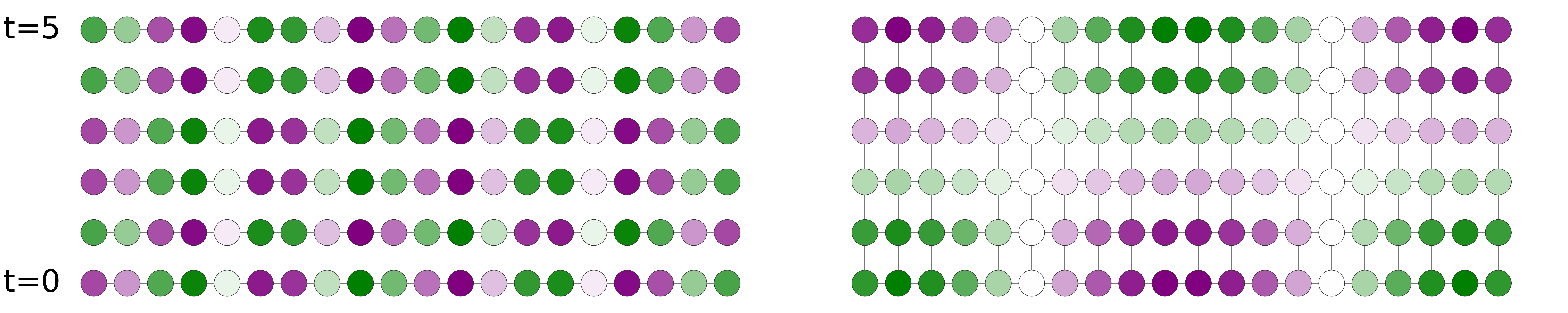}
    \caption{Comparison of eigenvector smoothness in multilayer single path graphs. Left: Graph without inter-layer connections exhibits less consistency. Right: Graph with inter-layer connections shows smooth eigenvector transitions, highlighting the role of inter-layer consistency.}
    \label{fig:eigenvectors-smoothness}
\end{figure}

The minimization in Equation \eqref{eq:smoothness_obj} shows that by using the supra-Laplacian eigenvectors, we achieve a balance between two key terms:
\begin{enumerate}
    \item \textit{Intra-layer smoothness}: The local connectivity structure of each layer \(G_t\) is preserved through the Laplacian quadratic form: \(\text{tr}\left({\mathbf{X}^{(t)}}^T \mathbf{L}_t \mathbf{X}^{(t)}\right)\).  Minimizing it makes the eigenvectors smooth according to the Laplacian \(\bfL_t\).
    \item \textit{Inter-layer consistency}: The penalty terms \(\mu \left\| \mathbf{X}^{(t)} - \mathbf{X}^{(t-1)} \right\|_F^2\), in contrast, enforces smooth transitions between eigenvectors of adjacent layers.
\end{enumerate}
The inter-layer consistency promoted by the smoothness term not only ensures the smoothness of inter-layer transitions but also encourages consistent sign assignments for eigenvectors, as can be seen in Figure \ref{fig:eigenvectors-smoothness}. This is in contrast to independently computed eigenvectors for each layer, where consecutive eigen-decompositions can lead to sign differences in each realization.

\subsection{The Expressiveness Benefits of Using the Supra-Adjacency}

Here, we shed light on the usefulness of considering the (multi-layer) supra-adjacency matrix rather than a single layer-wise approach. A key tool in our analysis is the Supra-Weisfeiler-Lehman (Supra-WL) test, which we define to extend the classical WL isomorphism test to snapshot-based temporal graphs represented as supra-graphs.
Supra-WL operates by iteratively refining node colors in the supra-graph: (i) it starts by assigning a constant color $\overline{c}$ to each node in the supra-graph $\mathbb{G}(\tau)$, or one which uniquely encodes node features, if available; (ii) it refines these colors by injectively hashing the current color, the colors of its temporal neighbors, as well as the multiset of colors of its spatial neighbors within the same layer $G(t)$:
\begin{equation}
    C_{v,t}^{(l+1)} = \text{HASH}\left(C_{v,t}^{(l)},C_{v,t-1}^{(l)},C_{v,t+1}^{(l)}, \multiset{(C_{u,t}^{(l)},e_{u,v,t},t)| (u,v,t) \in G(t)}\right)\label{eq:supra-wl-update} 
\end{equation}
\noindent where, for $t = 0$ and $t=\tau$ we have $C_{v,t-1}^{(l)} = C_{v,t+1}^{(l)} = \overline{\overline{c}}$.
The test is applied in parallel to two temporal graphs; it terminates when the multisets of node colors for the two supra-graphs diverge, indicating non-isomorphism. If the colors stabilize without divergence, the test is inconclusive.

To understand the importance of the information contained in the supra-adjacency, we compare the Supra-WL to Layer-WL, a simple extension of WL to snapshot-based temporal graphs. Layer-WL runs 1-WL color refinement steps independently on each graph snapshot $G(t)$, comparing the overall multisets of node colors thereon. We now present a critical distinction between the two algorithms, shedding light on the enhanced capabilities of the Supra-WL test and, by extension, models considering supra-adjacency information.

\begin{figure}[t]
    \centering
    \begin{minipage}[b]{0.48\linewidth}
        \includegraphics[width=\linewidth]{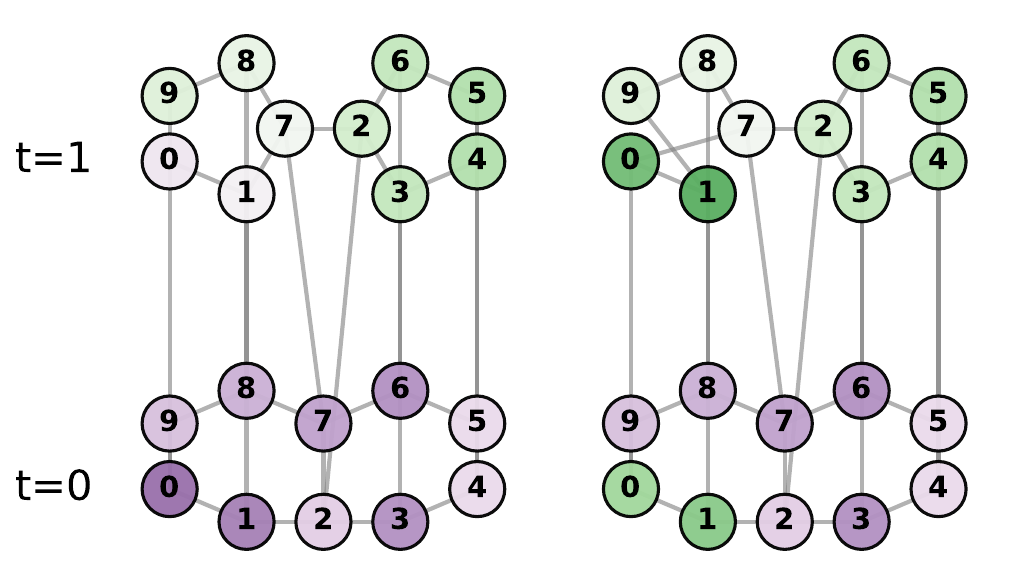}
    \end{minipage}
    \hfill
    \begin{minipage}[b]{0.48\linewidth}
        \includegraphics[width=\linewidth]{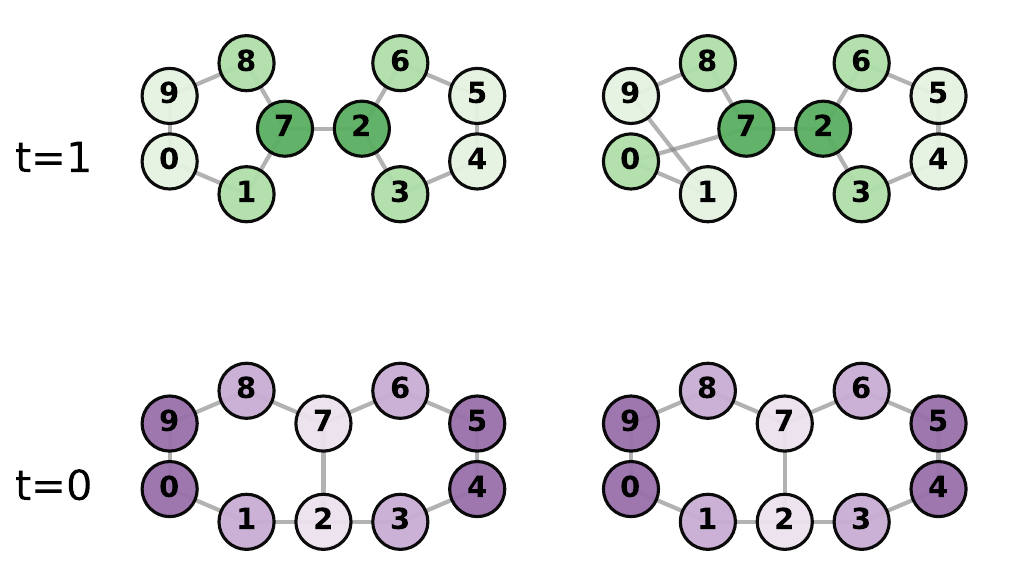}
    \end{minipage}
    \caption{Supra-WL correctly distinguishes the depicted non-isomorphic temporal graphs, while Layer-WL fails. As for the former, we explicitly depict the temporal connections induced by the supra-adjacency.}
    \label{fig:exampleA}
\end{figure}

\begin{proposition}  \label{prop:swl-vs-lwl-expressivity} (Supra-WL $\sqsubset$ Layer-WL). Supra-WL is
strictly more powerful than Layer-WL in distinguishing non-isomorphic DTDGs.
\end{proposition}
The proof of this proposition is deferred to Appendix \ref{proof:suprawl-vs-layerwl}, and involves exhibiting a pair on non-isomorphic DTDGs that are distinguished by Supra-WL but not by Layer-WL, reported in Fig~\ref{fig:exampleA}. 
This example emphasizes how 
treating layers as interconnected rather than independent is essential for capturing structural differences in temporal or multi-layer graphs that would otherwise go unnoticed.

\section{Efficient Computation of Supra-Laplacian PEs}\label{sec:efficient}
Focusing on enhancing the efficiency of eigendecompositions for the scalable computation of (S)LPEs, we propose and evaluate several strategies centered around iterative eigendecomposition and trajectory-based analysis. The overview of the proposed procedure is illustrated in Fig~\ref{fig:overview}.

First, we propose exploring the use of the Lanczos method~\cite{Lanczos:1950zz}, an iterative algorithm designed for large, sparse, symmetric matrices like graph Laplacians. It constructs a Krylov subspace and diagonalizes a smaller tridiagonal matrix, achieving exact solutions with sufficient iterations. 
Second, questioning the necessity of exact eigendecomposition, we propose to utilize solutions derived from the Locally Optimal Block Preconditioned Conjugate Gradient (LOBPCG) method~\cite{knyazev2017recentimplementationsapplicationsextensions}. LOBPCG is another iterative algorithm specifically engineered to efficiently compute a limited number of extreme eigenvalues and eigenvectors. LOBPCG offers memory efficiency, particularly when only a subset of eigenvectors is required. 

Furthermore, to leverage the information computed during iterative eigendecomposition solvers, we introduce a trajectory-based approach. Inspired by the work of \cite{10.5555/3618408.3618777} on static graphs, we extend this technique to temporal graphs. This approach recognizes that intermediate results from iterative solvers can be valuable and proposes to concatenate these intermediate results rather than solely relying on the achieved approximated solution. To address the inherent sign ambiguity of eigenvectors, for each eigenvector, we randomly choose a sign (+1 or -1) as was done in \cite{Dwivedi2020-ce} and consistently apply this sign across all iterations of its trajectory. Specifically, at each iteration $k$ of the eigendecomposition algorithm, let $\mathbf{U}^{(k)} \in \mathbb{R}^{n \times k}$ be the matrix of eigenvectors and $\mathbf{\Lambda}^{(k)}$ be the corresponding eigenvalues.  Our trajectory-based approach constructs concatenated representations as follows: $\mathbf{U}_{\text{traj}} = \left[ \mathbf{U}^{(1)}, \mathbf{U}^{(2)}, \dots, \mathbf{U}^{(K)} \right], \mathbf{\Lambda}_{\text{traj}} = \left[ \mathbf{\Lambda}^{(1)}, \mathbf{\Lambda}^{(2)}, \dots, \mathbf{\Lambda}^{(K)} \right]$.
Here, $K$ represents the total number of iterations, determined by either convergence criteria or a predefined early stopping point. This concatenated form aims to capture the evolution of eigenvectors and eigenvalues across iterations, potentially providing a richer representation of the temporal graph dynamics.

\section{Experiments}
In this section, we present a detailed evaluation of Laplacian-based PE variants for temporal graphs with various architectures. Our experiments aim to assess the impact of these approaches on downstream performance across diverse real-world datasets and feature configurations. In particular, our focus is on comparing standard graph representations, such as single snapshot-based graphs, with supra-graph representations, which combine multiple snapshots into a single supra-graph. Additionally, we investigate the effects of using iterative and approximate solvers for eigendecomposition in these models.
We aim to address the following questions:
\begin{enumerate}
    \item Do Laplacian-based PEs enhance the performance of TGNNs, in general?
    \item Which Laplacian-based PE scheme is best suited for TGNNs? 
    \item How do node features impact the performance of the Laplacian-based PEs?
    \item What are the computational benefits of \emph{approximate} Laplacian-based PEs?
\end{enumerate}
Full details on experiments and additional results are provided in Appendix~\ref{app:experimental-results}.

\subsection{Experimental Setup}
We evaluate four temporal graph models: EGCN~\cite{Pareja2019EvolveGCNEG}, GRUGCN~\cite{seo2017structured}, HTGN~\cite{Yang2021-rs}, and SLATE~\cite{karmim2024supralaplacian}. More details about these models can be found in Appendix~\ref{app:models}.
In our implementation of SLATE, we separated the SLPE from the architecture, allowing the investigation of different PE alternatives. In addition, we conducted a time analysis comparing the full eigendecomposition, Lanczos (That we run till convergence), and LOBPCG methods for computing the first 8 eigenvectors of both synthetic and real-world datasets. The overview of the proposed procedure is illustrated in Figure~\ref{fig:overview}.

\textbf{Node features.} We process the datasets under various feature configurations. We start with one-hot encodings, a common approach in the literature~\cite{Pareja2019EvolveGCNEG,10.5555/3455716.3455786}, that assigns a unique identifier to each node in the observed snapshot. However, we argue that this method may not be realistic for dynamic graphs where the number of nodes can grow unpredictably over time. Additionally, one-hot encodings may be suboptimal in the presence of nodes with few interactions, as the parameters associated with such nodes could remain undertrained and lead to poorer generalization performance.
Accordingly, we explore two additional feature encodings and their interplay with PEs. First, we experiment with uninformative, constant (zero) features, simulating the absence of node-specific information. Second, we study the impact of random node features. Although lacking any temporal and structural inductive bias, random features can, in principle, allow the model to identify nodes throughout its computations \cite{ACGL-IJCAI21}.

\textbf{Laplacian-based PE variants.} As a baseline, we report results without PEs (No PEs). In other cases, we compare several variants involving different PEs: SLPE and LPE, and different types of approximation: Exact eigenvalue computation \emph{(E)}, Inexact computation \emph{(I)}, and using the computation trajectory \emph{(T)}, as explained in Section~\ref{sec:efficient}. We use the Lanczos method for (E), appropriately run until convergence. LOBPCG is employed for (I). As for (T), we employ the trajectory generated by the latter. We note that the modification to the graph mentioned in Section~\ref{paragraph:slpe} and illustrated in Figure~\ref{fig:adj-matrix} are also done to the standard Laplacian.

\begin{table}[t!]
\caption{AUC performances of models across datasets and configurations for one-hot features. The top three models are highlighted by \space \first{First}, \space \second{Second}, \space \third{Third}.}
\label{tab:auc-one-hot-results}
\centering
\setlength{\tabcolsep}{2pt} 
\renewcommand{\arraystretch}{1.2} 
\begin{tabular}{llcccc}
\toprule
Dataset & Variant & EGCN & GRUGCN & HTGN & SLATE\\ \hline
CanParl  & No PEs & \first{85.56±0.27} & 67.10±1.54 & 87.59±0.69 & 56.22±1.31\\
         & SLPE-E   & 83.53±1.59 & \second{72.92±1.90} & \third{89.47±0.29} & \second{59.32±0.63}\\
         & LPE-E   & \second{85.41±1.44} & \first{74.92±0.74} & \first{89.62±0.16} & \first{59.99±1.07}\\
         & SLPE-I   & 83.45±1.59 & \third{72.71±1.41} & 89.26±0.55 & 57.42±0.97\\
         & LPE-I   & \third{84.18±1.06} & 71.53±0.73 & \second{89.61±0.19} & 56.96±0.69\\
         & SLPE-T   & 82.46±1.12 & 65.54±1.81 & 88.90±0.77 & \third{58.71±1.44}\\
         & LPE-T   & 81.72±0.89 & 67.11±1.79 & 88.98±0.31 & 55.04±1.28\\ \midrule
as733    & No PEs & 92.47±0.04 & 94.96±0.35 & \first{98.75±0.03} & \first{99.85±0.01}\\
         & SLPE-E   & \third{93.54±0.87} & 95.46±1.59 & \second{98.28±0.33} & 99.81±0.02\\
         & LPE-E   & \first{94.00±0.88} & \first{96.93±0.13} & \third{98.10±0.19} & \second{99.84±0.01}\\
         & SLPE-I   & \second{93.99±1.37} & 95.07±0.69 & 97.81±0.21 & \second{99.84±0.01}\\
         & LPE-I   & 93.52±0.65 & 96.13±1.11 & 97.61±0.34 & 99.80±0.01\\
         & SLPE-T   & 93.19±0.89 & \third{96.75±0.11} & 91.62±0.75 & 99.81±0.02\\
         & LPE-T   & 92.03±0.20 & \second{96.79±0.40} & 86.92±1.43 & 99.81±0.01\\ \midrule
dblp     & No PEs & 83.88±0.53 & 84.60±0.92 & \first{89.26±0.17} & 89.43±0.42\\
         & SLPE-E   & \first{87.10±0.23} & \second{86.93±0.96} & 88.67±0.55 & \first{89.68±0.56}\\
         & LPE-E   & 82.57±0.60 & 86.89±0.70 & \third{88.74±0.15} & 89.25±0.28\\
         & SLPE-I   & \second{86.66±0.34} & \second{86.93±0.46} & \second{88.77±0.12} & 89.38±0.42\\
         & LPE-I   & 83.90±0.48 & \first{87.04±0.98} & 88.52±0.36 & 89.40±0.19\\
         & SLPE-T   & \third{85.57±0.38} & 86.29±0.78 & 88.59±0.31 & \second{89.51±0.30}\\
         & LPE-T   & 80.67±0.60 & 86.70±0.62 & 88.08±0.37 & \third{89.46±0.34}\\ \midrule
enron10  & No PEs & 90.12±0.69 & 92.47±0.36 & 94.17±0.17 & \first{95.66±0.45}\\
         & SLPE-E   & \first{91.54±0.69} & \second{93.51±0.27} & \first{94.49±0.08} & \second{95.60±0.27}\\
         & LPE-E   & \third{90.48±0.64} & 93.34±0.83 & \third{94.37±0.24} & \third{95.49±0.33}\\
         & SLPE-I   & \second{91.40±0.75} & \first{93.63±0.13} & \second{94.45±0.54} & \third{95.49±0.31}\\
         & LPE-I   & 89.89±0.31 & \third{93.45±0.48} & \third{94.37±0.20} & \third{95.49±0.16}\\
         & SLPE-T   & 89.89±1.27 & 92.62±1.01 & 92.99±0.52 & 95.41±0.30\\
         & LPE-T   & 88.13±0.97 & 92.90±0.59 & 93.36±0.30 & 95.43±0.16\\\bottomrule
\end{tabular}
\end{table}

\textbf{Datasets, task, and performance metric.} The mentioned models are tested on the real-world datasets: CanParl, as733, dblp, and enron10~\cite{Yu2023-cv,Yang2021-rs}. Each represents dynamic graphs derived from snapshot-based observations (see Appendix~\ref{app:datasets} for more details). The task is (dynamic) link prediction in all cases. Performance is measured using the Area Under the Curve (AUC) metric, reported as the mean and standard deviation over five runs. The top three performing configurations for each model-dataset pair are highlighted as \space \first{First}, \space \second{Second}, and \space \third{Third}, respectively, based on the mean test AUC score.

The results are presented in Table~\ref{tab:auc-one-hot-results} (one-hot node features), with additional results provided in Table \ref{tab:auc-zeros-results} (using constant-zero node features), and Table \ref{tab:auc-randn-results} (using random node features) in the Appendix. Each table compares the performance of each of the aforementioned architectures across all datasets and PE schemes. The summarized results are presented in Table \ref{tab:avg-per-feat-model-with-pes} and Table \ref{tab:variant-and-features-averages}.

\begin{figure}[t]
    \centering
    \begin{minipage}{0.475\linewidth}
        \centering
        \includegraphics[width=\linewidth]{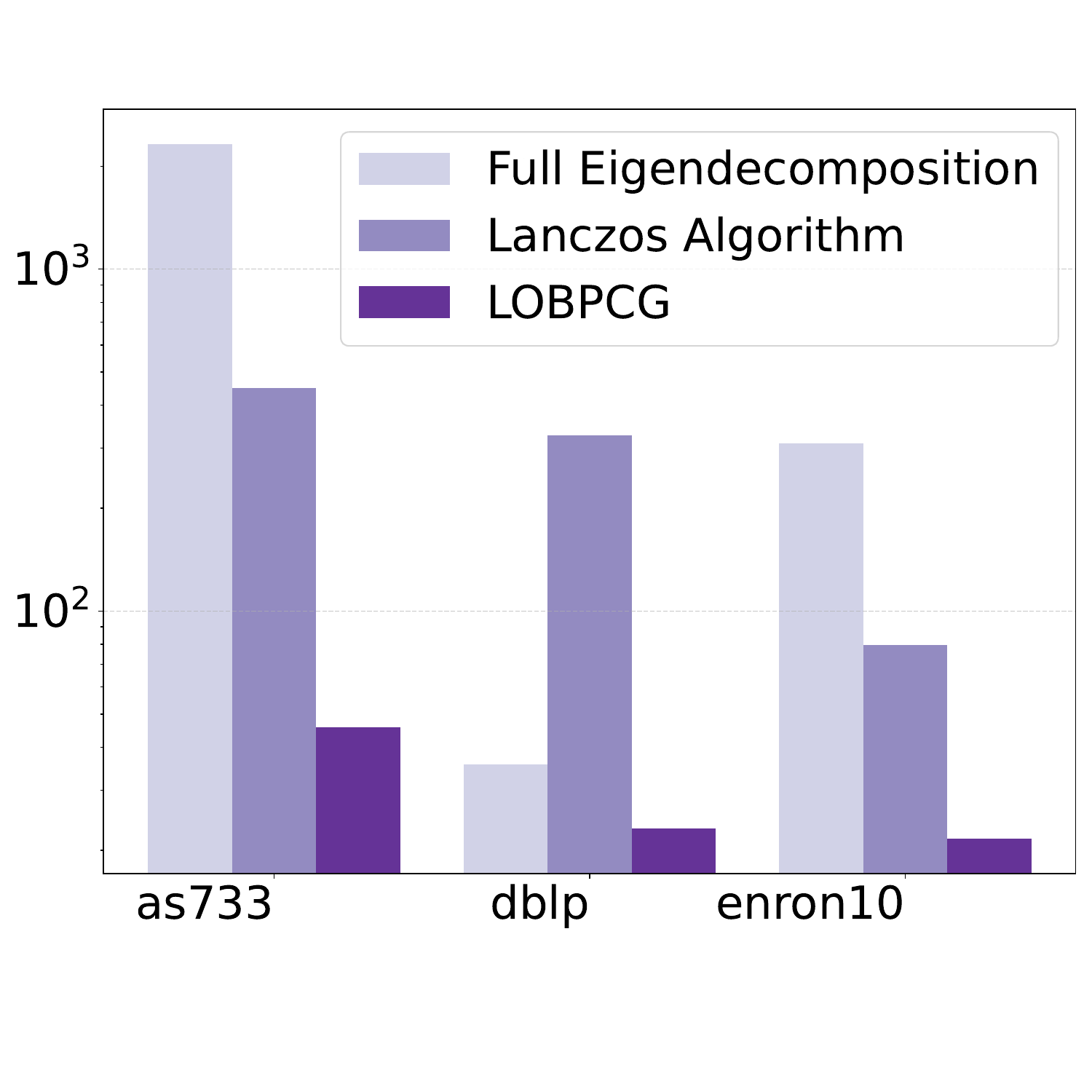}
    \end{minipage}
    \hfill
    \begin{minipage}{0.475\linewidth}
        \centering
        \includegraphics[width=\linewidth]{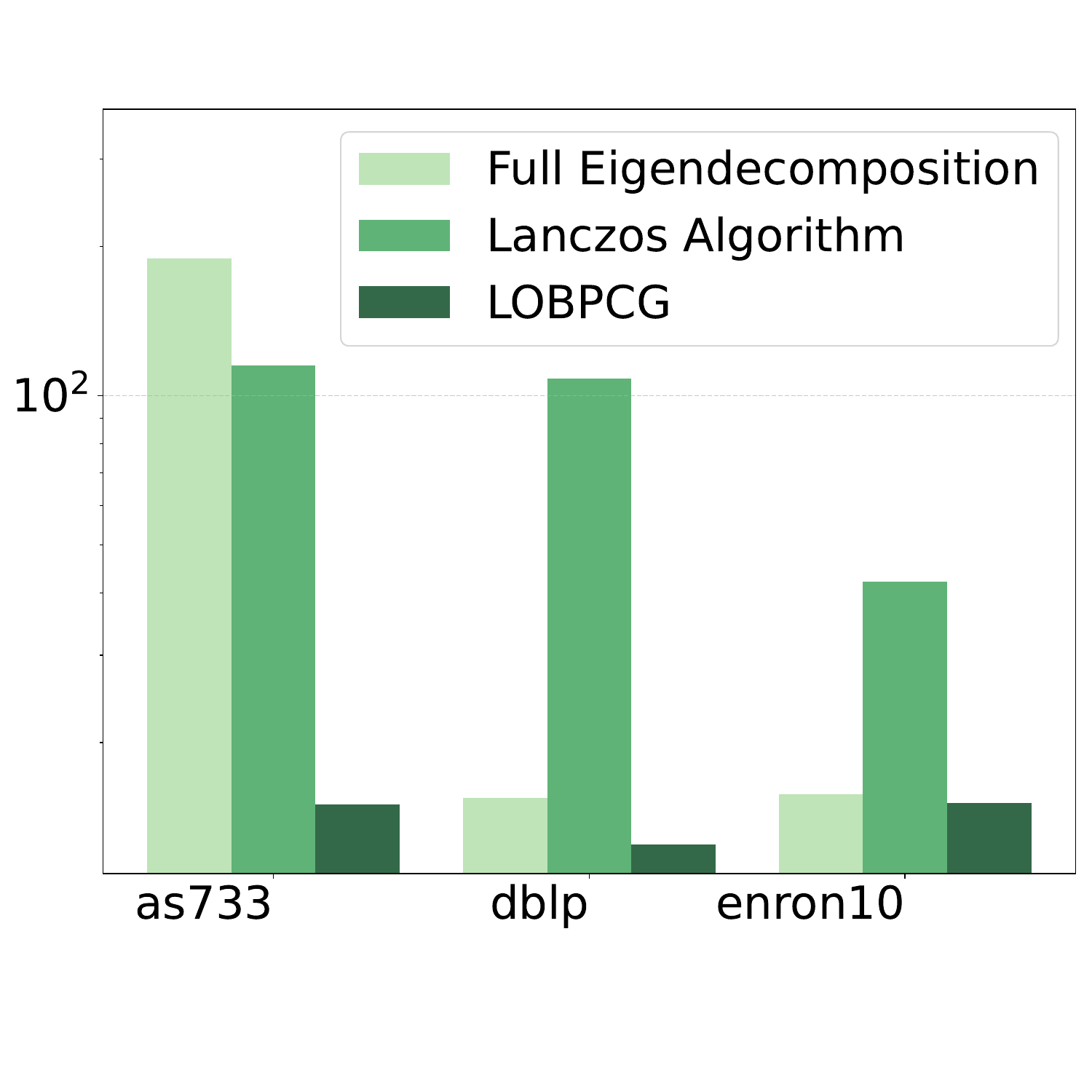}
    \end{minipage}
    \caption{Time (ms) performance comparison of Full Eigendecomposition, Lanczos, and LOBPCG methods on a real-world dataset presented by a Supra-Graph (left) and a single-layer-graph (right).}\label{fig:supra-regular-comparison}
\end{figure}

\subsection{Results and Discussion}

\textbf{Laplacian-based PEs \& TGNNs (Q1).} 
From Table~\ref{tab:auc-one-hot-results}, Table~\ref{tab:auc-zeros-results} and Table~\ref{tab:auc-randn-results} we observe that in $\approx 70.8\%$ of the overall number of cases, using Laplacian-based PEs led to the top-scoring results (\first{First}), and that in $\approx 64.5\%$ of the experiments, all the top-three ranking models employ Laplacian-based PEs. Quantitatively, Table \ref{tab:performance_diff} in Appendix \ref{app:additional-results} reports statistics on the absolute performance improvements induced by Laplacian-based PEs for each architecture. In all cases, except for HTGN, we observe positive median performance improvements, with the largest impact attained on SLATE ($16.63\%$) and EGCN ($8.20\%$). Importantly, we also observe how PEs are most useful when employing constant-zero node features, where they scored \space \first{First} in $\nicefrac{13}{16}$ cases, as shown in Table \ref{tab:auc-zeros-results}. As for the other feature configurations, they ranked\ \space \first{First} in \nicefrac{11}{16} cases (one-hot) and \nicefrac{8}{16} cases (random). From Table \ref{tab:diff-by-feature-model} in Appendix \ref{app:additional-results}, we note, more specifically, positive median AUC improvements are more pronounced in the case of constant features, where even HTGN seems to generally benefit from PEs. Quantitative improvements are also recorded, on average, for EGCN and GRUGCN across all feature variants, as well as SLATE when not using one-hot features. 
We conclude that Laplacian-based PEs are generally useful in improving generalization performance, and they are more consistent when the model is not provided with node-identifying information.

\textbf{Best Suited Laplacian-based PE variant for TGNNs (Q2).}
First, we compare SLPE variants to LPE variants. We found SLPE to perform better in 64.6\% of considered cases, with an aggregated average improvement of 1.09\% across models and datasets, as shown in Table \ref{app:results-diff-per-model}. The improvement is more consistent, in particular in EGCN and SLATE, where the average absolute improvements are, resp., $3.91\%$ and $0.66\%$. Notably, GRUGCN shows a small preference toward LPEs with an average performance difference of $0.30\%$. Next, we compare Exact variants with their Inexact counterparts. Overall, the former ones outperform the latter in 62.5\% of cases, but the improvement is less pronounced in this case. As can be observed in Table \ref{app:results-diff-per-model} (Appendix \ref{app:additional-results}), the distribution of performance differences is more dispersed, with median values close to zero in the case of HTGN and SLATE and only slightly in favor of Exact variants for EGCN and GRUGCN. This suggests the faster Inexact variants constitute promising candidates for more efficient pipelines. 

Finally, we compare variants overall, commenting on aggregated per-variant average performances as reported in Table \ref{app:results-avg-per-variant} in the aforementioned Appendix. Across settings, SLPE-I leads with an average AUC of 86.96\%, followed by SLPE-E with 86.38\%, further confirming the suitability of approximate, iterative eigensolvers.

\begin{table}[t]
    \centering
    \begin{minipage}[t]{0.55\textwidth} 
        \centering
        \caption{Average AUC (\%) performance per feature and model when \emph{Laplacian-based PEs are used} along with the difference between the max and min performance of different features for each model ($\Delta$).}\label{tab:avg-per-feat-model-with-pes}
        \begin{tabular}{lccccc}
        \toprule
        Feature & EGCN & GRUGCN & HTGN & SLATE & Avg \\
        \midrule
        one-hot & 87.87 & 86.75 & 91.73 & 85.66& 88.00 \\
        random & 81.72 & 80.89 & 88.59 & 77.63&  82.21\\
        constant & 87.16 & 84.74 & 89.48 & 80.55 & 85.48\\
        \midrule
        $\Delta$  & 6.15 & 5.86 & 3.14 & 8.03 &5.79\\
        \bottomrule
        \end{tabular}
    \end{minipage}
    \hfill
    \begin{minipage}[t]{0.4\textwidth}
        \centering
        \caption{Average AUC (\%) performances across Laplacian-based PEs and feature inits.}
        \label{tab:variant-and-features-averages}
        \begin{tabular}{lccc}
            \toprule
            Variant & one-hot & random & constant \\
            \midrule
            No PEs & 87.63 & 78.00 & 68.75 \\
            \hline
            LPE-E & \first{88.75} & 83.46 & 86.43 \\
            LPE-I & 88.21 & 80.96 & 85.0 \\
            LPE-T & 86.45 & 78.79 & 84.16 \\
            \hline
            SLPE-E & \second{88.74} & \second{83.57} & \second{86.83} \\
            SLPE-I & 88.52 & \first{85.14} & \first{87.23} \\
            SLPE-T & 87.37 & 81.33 & 83.24 \\
            \bottomrule
        \end{tabular}
    \end{minipage}
\end{table}

\textbf{Impact of features when coupled with Laplacian-based PEs (Q3).} Refer to Table \ref{tab:avg-per-feat-model-with-pes} for detailed per-model averages. When PEs are used, one-hot features outperform others with an aggregated average AUC of 88.00\%. They excel across models (e.g., 91.73\% for HTGN) and rank as the top performers in 18--23 cases per model. We argue that while one-hot encodings yield higher-performing models, those may be less practical for real-world scenarios where the number of nodes is unknown. Other features trail behind in both performance and frequency, with an average AUC of 85.48\% for constant (zero) features and 82.21\% for random features. A complementary angle to this discussion is offered by Table~\ref{tab:variant-and-features-averages}, where we report the average performance across models and datasets for each PE variant and feature choice. In agreement with our discussion in regards to Q1, we observe that PEs are most beneficial when using constant features. In addition to this, we note how the choice of PE variant is less impactful for one-hot features, while it leads to more result variability in the case of constant and random features. This effect is particularly pronounced in the latter case. In both settings, SLPEs achieve the best performance on average.
\begin{figure}[t]
    \centering
    \includegraphics[width=0.8\linewidth]{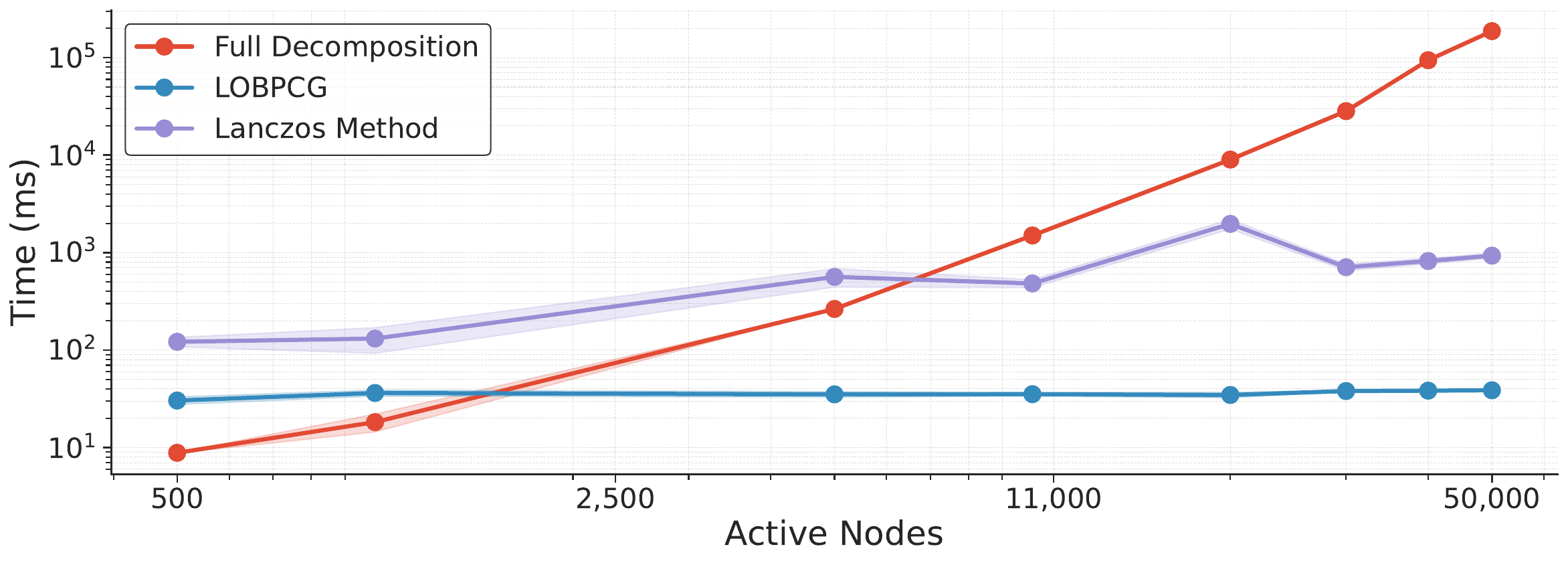}
    \caption{Time (ms) comparison of Full Eigendecomposition, Lanczos, and LOBPCG methods for computing the first 8 eigenvectors on Barabási–Albert graphs.}
    \label{fig:time-comparision-solvers}
\end{figure}

\textbf{Computational Benefits of approximate Laplacian-based PEs (Q4).}
Our time measurements, depicted in  Figure~\ref{fig:supra-regular-comparison}, show that LOBPCG is the fastest method for calculating the approximate Laplacian, consistently outperforming Lanczos and Full Eigendecomposition in both regular and Supra-graphs. In an effort to extend our time comparisons to even larger graphs, we synthesized random Barabási–Albert graphs with up to 50,000 active nodes, and timed the different methods thereon. See Figure~\ref{fig:time-comparision-solvers}. We observe that LOBPCG achieves a maximum speed-up of 56 times over Lanczos, with its efficiency advantage growing as the effective size of the graph increases.

\textbf{Results Summary.} 
In summary, our experiments demonstrate that integrating both LPEs and SLPEs into TGNNs generally enhances performance, especially when node features provide less discriminative information. When the node set is known in advance, one‐hot features lead to the best performance, while constant features remain a solid alternative when coupled with Laplacian-based PEs, especially SLPEs. Specifically, regarding the choice of PEs: the SLPEs variants generally outperform LPEs except in the case of HTGN. Most notably, the SLPE-I variant emerges as a robust default, balancing high accuracy (86.96\% average) with computational efficiency. Finally, our evaluation of eigenvector solvers reveals that approximate methods such as LOBPCG offer significant speed-ups, making them more suitable for large-scale graphs.

\section{Conclusions}
In this paper, we have thoroughly reviewed Laplacian-based PEs within the framework of TGNNs, providing a comprehensive understanding of their role and limitations. Our theoretical analysis of SLPEs reveals significant insights into their expressive power and connection to single-layer Laplacian PEs. In addition, we have demonstrated the practical implications of various PEs, providing actionable guidelines for practitioners seeking to optimize TGNN performance. Our findings highlight the benefits of incorporating Laplacian-based PEs and their interplay with node feature initialization schemes, underscoring how faster, approximate eigendecompositions can maintain a compelling tradeoff between run-time and model performance. We believe this work opens up interesting future research directions. These include exploring the study of further, more efficient eigensolvers for large graphs and the use of SLPEs in CTDGs.
\begin{credits}
\subsubsection{\ackname} F.F. conducted this work supported by an Andrew and Erna Finci Viterbi Fellowship and, partly, by an Aly Kaufman Post-Doctoral Fellowship. F.F. partly performed this work while visiting the Machine Learning Research Unit at TU Wien, led by Prof. Thomas Gärtner.
H.M. is the Robert J. Shillman Fellow, and is supported by the Israel Science Foundation through a personal grant (ISF 264/23) and an equipment grant (ISF 532/23). M.E. is funded by the Blavatnik-Cambridge fellowship, the Cambridge Accelerate Programme for Scientific Discovery, and the Maths4DL EPSRC Programme. E.T. was partially supported by the Israeli Council for Higher Education (CHE) via Data Science Research Center, Ben-Gurion University of the Negev, Israel. 

\subsubsection{\discintname} The author declares no competing interests relevant to the content of this article.
\end{credits}

\bibliographystyle{splncs04}
\bibliography{paper}

\newpage
\clearpage
\title{Supplementary Material: Understanding and Improving Laplacian Positional Encodings For Temporal GNNs}
\author{}
\titlerunning{Supplementary Material}
\authorrunning{Y. Galron et al.}
\institute{}
\maketitle
\appendix 
\section{Datasets} \label{app:datasets}
Datasets statistics are presented in Table~\ref{tab:datasets}
\begin{table}[t]
\captionof{table}{Dataset statistics used in our experiments from \cite{Yu2023-cv} and  \cite{Yang2021-rs}.}
  \label{tab:datasets}
  \centering
  \begin{tabular}{lllll} 
    \toprule
    Datasets & Domains & Nodes & Links & Snapshots \\
    \midrule
    CanParl & Politics & 734 & 74,478 & 14 \\
    \midrule
    AS733 & Router & 6,628  & 13,512 & 30 \\
    \midrule
    Enron & Mail & 184 & 790  & 11 \\
    \midrule
    dblp & Citations & 315 & 943 & 10 \\
    \bottomrule
  \end{tabular}
\end{table}

\begin{itemize}
    \item \textbf{CanParl}: Can. Parl. is a network that tracks how Canadian Members of Parliament (MPs) interacted between 2006 and 2019. Each dot represents an MP, and a line connects them if they both said "yes" to a bill. The line's thickness shows how often one MP supported another with "yes" votes in a year.
    \item \textbf{Enron}: Enron consists of emails exchanged among 184 Enron employees. Nodes represent employees, and edges indicate email interactions between them. The dataset includes 10 snapshots and does not provide node or edge-specific information
    \item \textbf{dblp}: dblp represents an academic cooperation network, capturing the collaborative efforts of 315 researchers from 2000 to 2009. In this network, each node corresponds to an author, and an edge signifies a co-authorship relationship.
    \item \textbf{AS733}: AS733 represents an Internet router network dataset, compiled from the University of Oregon Route Views Project. This dataset consists of 733 instances, covering the time period from November 8, 1997, to January 2, 2000, with intervals of 785 days between data points.
\end{itemize}

\subsection{Datasets split}

For the datasets from \cite{Yu2023-cv}, we follow the same graph splitting strategy, which means 70\% of the snapshots for training, 15\% for validation, and 15\% for testing. 
We use the same number of snapshots as in HTGN \cite{Yang2021-rs}, the value varies for each dataset (Table \ref{tab:splitdtdg}).

\begin{table}[!hbt]
    \caption{$l$ represents the number of snapshots in the test dataset. The DTDG is split temporally, following \cite{Yang2021-rs}}
    \label{tab:splitdtdg}
    \centering
    \begin{tabular}{c|cccc}
    \toprule
        Datasets  & AS733  & Enron & Colab \\
        \midrule
        $l$ (number snapshots in \textit{test})  & 10 & 3 & 3\\
        \bottomrule
    \end{tabular}
    
\end{table}

\section{Models} \label{app:models}
\begin{itemize}
    \item \textbf{GRUGCN} \cite{seo2017structured}: GRUGCN is one of the first discrete dynamic graph GNN models. They introduced the now standard approach which combines a GNN to process the snapshot, and updating embeddings using a temporal model, in their case a GRU.
    \item \textbf{EvolveGCN} \cite{Pareja2019EvolveGCNEG}: EvolveGCN is an innovative approach adapting the Graph Convolutional Network (GCN) model for dynamically evolving graphs without relying on node embeddings, effectively capturing the dynamic nature of graph sequences through a Recurrent Neural Network to update GCN parameters.
    \item \textbf{HTGN} \cite{Yang2021-rs}: They introduce a novel approach for embedding temporal networks through a hyperbolic temporal graph network (HTGN), effectively utilizing hyperbolic space to capture complex, evolving relationships and hierarchical structures in temporal networks. 
    \item \textbf{SLATE} \cite{karmim2024supralaplacian}:SLATE transforms snapshot-based graphs into multi-layer graphs and leverages the spectral properties of the supra-Laplacian matrix to calculate PEs. Then a fully connected transformer is used, enabling accurate edge representation for dynamic link prediction. Our implementation of SLATE decouples the SLPE from the architecture and allows the incorporation of other PE alternatives.
\end{itemize}

\section{Proofs}

\subsection{Supra-Laplacian Layer-wise Eigenvectors Smoothness} \label{app:supra-smoothness-proof}
\begin{proof} 
First, let us define $\bfX$ as the concatenation of all temporal PE matrices $\mathbf{X}^{(t)}$ for a given snapshot: 
\begin{equation}\label{eq:X}
\mathbf{X} =
\begin{pmatrix} 
        \bfX^{(1)}\\
        \bfX^{(2)}\\
        \vdots \\
        \bfX^{(T-1)}\\
        \bfX^{(T)}
\end{pmatrix}.
\end{equation}
Next, let us write the non-circular Supra-Adjacency, Supra-Degree, and Supra-Laplacian matrices in terms of each specific layer's adjacency, degree, and Laplacian matrices. Here we consider one-to-one interconnections are one-to-one, and the strength of those connections is uniform and controlled by $\mu$. The supra-Adjacency is given by
\begin{equation}
\bfA_{\text{supra}} =
\begin{pmatrix} 
        \bfA_{1}&\mu \bfI&&&\\
        \mu \bfI&\bfA_{2}&\mu \bfI&&&\\
        &\mu \bfI&\bfA_{3}&\ddots&\\
        &&\ddots&\ddots&\mu \bfI\\
        &&&\mu \bfI&\bfA_{L}
\end{pmatrix}.
\end{equation}
The supra-degree matrix is given by
\begin{equation}
\bfD_{\text{supra}} =\begin{pmatrix} 
                        \bfD_1+\mu \bfI\\
                        &\bfD_2+2\mu \bfI&&\\
                        &&\ddots\\
                        &&&\bfD_{T-1}+2\mu \bfI\\
                        &&&&\bfD_T+\mu \bfI
                        \end{pmatrix},
\end{equation}
and the supra-Laplacian is given by
\begin{equation}\label{eq:supra_Lap_explicit}
\bfL_{\text{supra}} =\bfD_{\text{supra}} - \bfA_{\text{supra}} =\begin{pmatrix} 
        \bfL_1+\mu \bfI&-\mu \bfI&&&\\
        -\mu\bfI&\bfL_{2}+2\mu \bfI&-\mu \bfI&&&\\
        &-\mu \bfI&\bfL_{3}+2\mu \bfI&-\mu \bfI&\\
        && \ddots&\ddots&\ddots\\
        &&&\ddots&\bfL_{T-1}+2\mu \bfI&-\mu \bfI\\
        &&&&-\mu \bfI&\bfL_{T}+\mu \bfI
\end{pmatrix}.
\end{equation}
Since the supra-Laplacian $\bfL_{\text{supra}}$ is a symmetric positive semi-definite matrix, it is known that the $d$ lowest eigenvectors of the supra-Laplacian can be computed as the following minimization problem:
\begin{equation}\label{eq:min_eigs}
\min_{\mathbf{X}^{T}\bfX = \bfI}  \text{tr}\left({\mathbf{X}}^T \mathbf{L}_{\text{supra}} \mathbf{X}\right).
\end{equation}
We now show that this minimization is equivalent to the minimization of \eqref{eq:smoothness_obj}. Using the definitions in \eqref{eq:X} and \eqref{eq:supra_Lap_explicit} and reading the matrix line by line, we have that
\begin{eqnarray}
\text{tr}\left({\mathbf{X}}^T \mathbf{L}_{\text{supra}} \mathbf{X}\right) &=& \sum_{t=1}^{T}{\text{tr}\left({\bfX^{(t)}}^T\bfL_{t}\bfX^{(t)}\right)} \\ \nonumber
&&-\mu\sum_{t=2}^{T-1} \text{tr}\left({\bfX^{(t-1)}}^T \bfX^{(t)} -2{\bfX^{(t)}}^T \bfX^{(t)} +{\bfX^{(t)}}^T \bfX^{(t+1)} \right)\\ \nonumber
&&+\mu\text{tr}\left({\bfX^{(1)}}^T \bfX^{(1)} - {\bfX^{(1)}}^T \bfX^{(2)} - {\bfX^{(T)}}^T \bfX^{(T-1)} + {\bfX^{(T)}}^T \bfX^{(T)} \right).
\end{eqnarray}
Rearranging the indices in the sums, we can write
\begin{eqnarray}
\text{tr}\left({\mathbf{X}}^T \mathbf{L}_{\text{supra}} \mathbf{X}\right) &=& \sum_{t=1}^{T}{\text{tr}\left({\bfX^{(t)}}^T\bfL_{t}\bfX^{(t)}\right)} \\ \nonumber
&&+\mu\sum_{t=2}^{T} \text{tr}\left({\bfX^{(t-1)}}^T \bfX^{(t-1)} - {\bfX^{(t-1)}}^T \bfX^{(t)} - {\bfX^{(t)}}^T \bfX^{(t-1)}+ {\bfX^{(t)}}^T \bfX^{(t)} \right)\\ \nonumber
&=& \sum_{t=1}^{T}{\text{tr}\left({\bfX^{(t)}}^T\bfL_{t}\bfX^{(t)}\right)}  +\mu\sum_{t=2}^{T} \text{tr}\left((\bfX^{(t)}-\bfX^{(t-1)})^T (\bfX^{(t)}-\bfX^{(t-1)}) \right)\\\nonumber
&=& \sum_{t=1}^T \text{tr}\left({\mathbf{X}^{(t)}}^T \mathbf{L}_t \mathbf{X}^{(t)}\right) + \mu \sum_{t=2}^T\left\| \mathbf{X}^{(t)} - \mathbf{X}^{(t-1)} \right\|_F^2.
\end{eqnarray}
The last equality is exactly the objective in \eqref{eq:smoothness_obj} and since the constraints in both problems is the orthogonality of the columns in $\bfX$ we essentially have the same problem in Equations \eqref{eq:smoothness_obj} and \eqref{eq:min_eigs}. 
\hfill$\blacksquare$
\end{proof}

\subsection{Supra-WL vs Layer-WL} \label{proof:suprawl-vs-layerwl}
We have previously demonstrated with Example \ref{fig:exampleA} that there exist graph instances where Layer-WL (LWL) fails to distinguish between graphs, while Supra-WL (SWL) succeeds.  Now, we aim to formally prove that if SWL assigns identical colors to two nodes, then LWL will also assign identical colors to the same nodes. This conclusion will eventually establish that SWL is strictly more expressive than LWL.
\begin{proof}
Recall the definition of LWL's color update rule, given a constant initial color $C_{v,t}^{(0)} = \overline{c}$ for all nodes:
\begin{equation}
    C_{v,t}^{(l+1)} = \text{HASH}\left(C_{v,t}^{(l)}, \multiset{\left(C_{u,t}^{(l)},e_{u,v,t},t\right)| (u,v,t) \in G(t)}\right)
\end{equation}
This rule highlights that LWL considers only nodes and edges within the current timestamp $t$.
Where $0\leq t\leq\tau$ .

For completeness, we restate the SWL update rule from equation \eqref{eq:supra-wl-update}:
\begin{equation}
C_{v,t}^{(l+1)} = \text{HASH}\left(C_{v,t}^{(l)},C_{v,t-1}^{(l)},C_{v,t+1}^{(l)}, \multiset{(C_{u,t}^{(l)},e_{u,v,t},t)| (u,v,t) \in G(t)}\right) 
\end{equation}
Where for $t = 0$ and $t= \tau$ we will have $C_{v,t-1}^{(l)}= c$  and  $C_{v,t+1}^{(l)}= c$  respectively. 

Supra-WL, in contrast, incorporates information from the same node in the previous ($t-1$) and next ($t+1$) timestamps, in addition to the current timestamp $t$.

We want to prove the following implication:
Let $c_{v,t}^{l}$ denote the color of node $v$ at timestamp $t$ and iteration $l$ for SWL, and $d_{v,t}^{l}$ denote the color for LWL. We aim to show that:
\begin{equation}
    c_{v,t}^{l} = c_{w,t}^{l} \implies d_{v,t}^{l} = d_{w,t}^{l} , \quad \forall t,v,w, l
\end{equation}
This implication will demonstrate that if SWL cannot distinguish nodes $v$ and $w$ (i.e., assigns them the same color), then LWL also cannot distinguish them.  Consequently, the coloring induced by SWL refines that of LWL ($SWL \sqsubseteq LWL$).

We will prove this statement using induction on the iteration level $l$.

\textbf{Base Case:} $l=0$.  Initially, all nodes are assigned the same color $\overline{c}$ for both SWL and LWL. Thus, if $c_{v,t}^{(0)} = c_{w,t}^{(0)} = \overline{c}$, then it trivially follows that $d_{v,t}^{(0)} = d_{w,t}^{(0)} = \overline{c}$. The base case holds.
\begin{equation}
    c_{v,t}^{(0)} = c_{w,t}^{(0)} = d_{v,t}^{(0)} = d_{w,t}^{(0)} = \overline{c}
\end{equation}

\textbf{Inductive Step:} Assume that the implication holds for iteration $l$.  That is, assume:
\begin{equation}\label{H_INDUCTION}
    c_{v,t}^{l} = c_{w,t}^{l} \implies d_{v,t}^{l} = d_{w,t}^{l}
\end{equation}
We will now prove that the implication holds for iteration $l+1$:
\begin{equation}
    c_{v,t}^{l+1} = c_{w,t}^{l+1} \implies d_{v,t}^{l+1} = d_{w,t}^{l+1}
\end{equation}

Suppose $c_{v,t}^{(l+1)} = c_{w,t}^{(l+1)}$.  According to the SWL update rule, this means:
\begin{equation}
\begin{split}
                \text{HASH}\left(c_{v,t}^{(l)},c_{v,t-1}^{(l)},c_{v,t+1}^{(l)}, \multiset{\left(c_{u,t}^{(l)},e_{u,v,t},t\right)| (u,v,t) \in G(t)}\right) = \\ \text{HASH}\left(c_{w,t}^{(l)},c_{w,t-1}^{(l)},c_{w,t+1}^{(l)}, \multiset{\left(c_{u',t}^{(l)},e_{u',w,t},t\right)| (u',w,t) \in G(t)}\right)
    \end{split}
\end{equation}
Since HASH is an injective function, equality of the hash outputs implies equality of their inputs. Therefore, we must have:
\begin{equation}\label{Supra_equals_current}
    c_{v,t}^{(l)} = c_{w,t}^{(l)}
\end{equation}
\begin{equation}\label{Supra_equals_prev}
    c_{v,t-1}^{(l)} = c_{w,t-1}^{(l)}
\end{equation}
\begin{equation}\label{Supra_equals_next}
c_{v,t+1}^{(l)}=c_{w,t+1}^{(l)}
\end{equation}
\begin{equation} \label{multiset_eq}
\multiset{\left(c_{u,t}^{(l)},e_{u,v,t},t\right)| (u,v,t) \in G(t)} = \multiset{\left(c_{u',t}^{(l)},e_{u',w,t},t\right)| (u',w,t) \in G(t)}
\end{equation}
In addition, from the inductive hypothesis \eqref{H_INDUCTION}, we know that:
\begin{equation}\label{Layer_equals_current}
    d_{v,t}^{(l)} = d_{w,t}^{(l)}
\end{equation}
And, if we apply the inductive hypothesis \eqref{H_INDUCTION} to the multisets in \eqref{multiset_eq}, we can replace the Supra-WL colors in the multisets with their corresponding Layer-WL colors without affecting the equality of the multisets resulting in:
\begin{equation} \label{multiset_eq_layer}
\multiset{\left(d_{u,t}^{(l)},e_{u,v,t},t\right)| (u,v,t) \in G(t)} = \multiset{\left(d_{u',t}^{(l)},e_{u',w,t},t\right)| (u',w,t) \in G(t)}
\end{equation}
Now, considering the LWL update rule for nodes $v$ and $w$ at timestamp $t$:
\begin{equation}
    d_{v,t}^{(l+1)} = \text{HASH}\left(d_{v,t}^{(l)}, \multiset{\left(d_{u,t}^{(l)},e_{u,v,t},t\right)| (u,v,t) \in G(t)}\right)
\end{equation}
\begin{equation}
    d_{w,t}^{(l+1)} = \text{HASH}\left(d_{w,t}^{(l)}, \multiset{\left(d_{u',t}^{(l)},e_{u',w,t},t\right)| (u',w,t) \in G(t)}\right)
\end{equation}
From \eqref{Layer_equals_current} we have $d_{v,t}^{(l)} = d_{w,t}^{(l)}$, and from \eqref{multiset_eq_layer} we have the equality of the multisets.  Therefore, the inputs to the HASH function are identical for both $d_{v,\tau}^{(l+1)}$ and $d_{w,\tau}^{(l+1)}$.  We conclude:
\begin{equation}
    d_{v,\tau}^{(l+1)} = d_{w,\tau}^{(l+1)}
\end{equation}
This completes the inductive step.

By induction, we have proven that for all iterations $l$, if $c_{v,t}^{l} = c_{w,t}^{l}$, then $d_{v,t}^{l} = d_{w,t}^{l}$.  This with combination with Example \ref{fig:exampleA} demonstrates that LWL strictly weaker than SWL in terms of distinguishing power, formally expressed as $SWL \sqsubset LWL$.

\hfill$\blacksquare$
\end{proof}
\section{Experimental Results} \label{app:experimental-results}
\subsection{Experimental settings} \label{app:exp-settings}
Our code is implemented using PyTorch~\cite{paszke2019pytorchimperativestylehighperformance} and PyTorch-Geometric~\cite{fey2019fastgraphrepresentationlearning}, and all our
experiments are run on Nvidia A100 GPUs with 40GB of memory.

\textbf{Hyperparameters}
We now list the hyperparameters used in our experiments. The learning rate is denoted by $lr$, weight decay by $wd$, and dropout probability by $drop$. The number of layers is denoted by $L$, and the number of hidden channels by $c$. Additionally, the Supra-Laplacian PES requires two main hyperparameters: the number of timestamps to consider (window size) $ws$, and the number of eigenvectors to be estimated $k$. The hyperparameters were determined using a Bayesian search, and the considered values are as follows:

\begin{itemize}
    \item Learning rate ($lr$): $\{10^{-1}, 10^{-2}, 10^{-3}, 10^{-4}\}$
    \item Weight decay ($wd$): $\{10^{-8}, 10^{-7}, 10^{-6}, 10^{-5}, 10^{-4}, 10^{-3}\}$
    \item Dropout probability ($drop$): Uniformly sampled from $[0.0, 0.5]$
    \item Number of layers ($L$): $\{2, 4, 8, 16\}$
    \item Number of hidden channels ($c$): $\{8, 16, 32, 64, 128\}$
    \item Window size ($ws$): Integer values uniformly sampled from $[2, 5]$
    \item Number of eigenvectors ($k$): Integer values uniformly sampled from $[4, 16]$
    \item PES initialization: $\{\text{normal}, \text{rademacher}, \text{uniform}, \text{with\_old\_pes}\}$
    \item Maximum iterations ($maxiter$): $\{5, 10, 20, 50\}$
\end{itemize}

In addition, for each architecture used, the additional specific hyperparameters were combined into the sweep arguments. Each experiment is repeated 5 times for robustness.
\newpage 
\subsection{Additional Results} \label{app:additional-results}
We present additional results with Area under the curve (AUC) metrics to evaluate the dynamic link prediction. The following tables offer summarized views into the behavior of different models across PE variants, features, and datasets.

\begin{table}[ht!]
\caption{AUC Performance comparison of models across datasets and configurations for zeros features initialization. The top three models are highlighted by \space \first{First}, \space \second{Second}, \space \third{Third}.}
\label{tab:auc-zeros-results}
\centering
\setlength{\tabcolsep}{2pt} 
\renewcommand{\arraystretch}{1.2} 
\begin{tabular}{llcccc}
\toprule
Dataset & Variant & EGCN & GRUGCN & HTGN & SLATE\\ \hline
CanParl  & No PEs   & 50.00±1.15  & 64.98±0.23  & 74.18±1.29  & 56.06±1.36\\
               & SLPE-E     & \third{84.15±0.67}  & \third{68.83±0.73}  & 85.19±0.57  & 58.14±1.17\\
               & LPE-E     & \second{84.41±1.87} & \first{80.94±0.71}  & 84.48±0.83  & \first{62.61±0.57}\\
               & SLPE-I     & \first{85.69±1.33}  & \second{71.07±0.72} & \second{87.60±0.58} & 57.93±0.72\\
               & LPE-I     & 83.49±0.94  & 66.83±0.64  & 84.67±0.89  & \third{59.14±0.84}\\
               & SLPE-T     & 81.70±1.69  & 66.86±1.77  & \third{87.49±1.51} & 58.37±1.23\\
               & LPE-T     & 82.59±0.98  & 64.99±0.62  & \first{87.91±0.68} & \second{60.23±1.09}\\ \midrule
as733   & No PEs   & 50.00±0.53  & 92.46±0.88  & 84.49±1.47  & 58.61±1.36\\
               & SLPE-E     & 94.07±1.01  & 93.92±1.20  & \second{90.01±1.82} & \first{97.22±0.11}\\
               & LPE-E     & 92.87±0.10  & \first{96.49±1.01} & \first{92.09±1.54} & 96.40±0.26\\
               & SLPE-I     & \first{94.96±0.72}  & 95.30±0.89  & 89.52±1.20  & \second{97.07±0.10}\\
               & LPE-I     & 92.67±0.08  & 93.17±1.49  & 89.53±0.66  & \third{96.60±0.44}\\
               & SLPE-T     & \second{94.33±0.15} & \third{95.38±0.97} & 88.86±1.46  & 96.39±0.27\\
               & LPE-T     & \third{94.29±0.30}  & \second{96.26±0.26} & \third{89.82±0.84} & 94.47±0.91\\ \midrule
dblp    & No PEs   & 50.00±1.11  & \first{86.14±0.55}  & \first{88.65±0.54}  & 52.67±1.50\\
               & SLPE-E     & \first{87.16±0.12}  & \second{85.97±0.64}  & \third{88.09±0.46}  & \first{86.70±0.71}\\
               & LPE-E     & 78.12±0.72  & 85.39±1.36  & 88.00±0.38  & 79.84±0.53\\
               & SLPE-I     & \second{86.78±0.34} & \third{85.86±0.76} & \second{88.18±0.31} & \second{86.16±1.01}\\
               & LPE-I     & 78.73±0.07  & 85.23±1.05  & 88.07±0.58  & 80.27±0.54\\
               & SLPE-T     & \third{85.44±0.51}  & 80.07±1.53  & 87.93±0.76  & \third{81.97±1.61}\\
               & LPE-T     & 79.58±0.36  & 72.20±0.93  & \third{88.09±0.21}  & 77.55±0.52\\ \midrule
enron10 & No PEs   & 50.00±1.44  & \second{93.00±0.75}  & \first{94.15±0.11}  & 54.62±1.88\\
               & SLPE-E     & \third{89.78±0.52}  & 92.88±0.51  & \second{93.84±0.33}  & \first{93.39±0.34}\\
               & LPE-E     & 86.05±1.62  & \first{93.02±0.31}  & 93.62±0.37  & 88.48±0.61\\
               & SLPE-I     & \first{90.78±0.66}  & 92.41±0.55  & \third{93.76±0.38}  & \second{92.56±0.90}\\
               & LPE-I     & 85.93±0.82  & \first{93.02±0.96}  & 93.70±0.46  & \third{88.90±0.70}\\
               & SLPE-T     & \second{90.21±0.92} & 87.86±1.29  & 93.48±0.67  & 55.47±0.96\\
               & LPE-T     & 87.96±0.82  & 89.87±0.75  & 93.47±0.69  & 87.22±0.23\\ \midrule
\end{tabular}
\end{table}

\begin{table}[ht!]
\caption{AUC Performance comparison of models across datasets and configurations for random normal features.
The top three models are highlighted by \space \first{First}, \space \second{Second}, \space \third{Third}.}
\label{tab:auc-randn-results}
\centering
\setlength{\tabcolsep}{2pt} 
\renewcommand{\arraystretch}{1.2} 
\begin{tabular}{llcccc}
\toprule
Dataset & Variant & EGCN & GRUGCN & HTGN & SLATE\\ \hline
CanParl  & No PEs   & \first{84.25±0.52}  & 64.96±1.49  & \first{83.95±1.74}  & \second{58.92±1.01}\\
               & SLPE-E     & \second{82.92±0.41} & \second{69.44±1.52} & \second{83.73±0.51} & 54.87±1.15\\
               & LPE-E     & \third{82.61±0.51}  & \first{75.57±1.31}  & 82.98±1.32  & \first{59.05±0.53}\\
               & SLPE-I     & 82.36±0.43  & 64.67±0.72  & 83.32±0.84  & 54.17±0.53\\
               & LPE-I     & 55.56±1.31  & \third{68.40±1.04} & \third{83.69±1.59} & \third{58.90±1.09}\\
               & SLPE-T     & 82.00±0.50  & 66.85±1.00  & 82.80±1.11  & 58.29±0.72\\
               & LPE-T     & 81.42±0.57  & 64.78±1.07  & 80.87±1.30  & 56.03±0.70\\ \midrule
as733   & No PEs   & \third{83.20±0.78}  & 92.72±0.85  & \first{97.36±0.17}  & 63.12±1.05\\
               & SLPE-E     & 75.48±1.39  & 93.70±1.80  & 89.74±1.23  & \second{95.78±0.36}\\
               & LPE-E     & 73.81±0.92  & \second{94.23±0.54} & \third{92.80±1.35}  & 94.64±0.24\\
               & SLPE-I     & \first{94.95±0.89}  & \first{94.98±0.25}  & \second{93.09±1.31}  & \first{96.07±0.93}\\
               & LPE-I     & 74.95±0.59  & \third{94.16±1.08} & 91.29±1.04  & \third{95.38±0.24}\\
               & SLPE-T     & \second{85.42±1.18} & 93.74±0.71  & 84.51±0.60  & 64.33±1.33\\
               & LPE-T     & 73.64±1.24  & 93.72±0.59  & 87.95±1.10  & 91.89±0.85\\ \midrule
dblp    & No PEs   & 74.04±1.36  & 68.87±1.08  & \first{88.65±0.68}  & 58.87±0.58\\
               & SLPE-E     & \first{83.92±1.45}  & \first{74.51±0.77}  & \second{88.22±0.58}  & \second{83.60±0.92}\\
               & LPE-E     & 77.56±0.97  & \third{73.87±0.86}  & 87.80±0.35  & 78.67±0.48\\
               & SLPE-I     & \second{83.42±0.64} & 73.78±0.71  & 87.70±0.50  & \first{85.58±0.57}\\
               & LPE-I     & 77.93±0.50  & 73.74±0.78  & \third{87.86±0.38}  & 78.55±0.73\\
               & SLPE-T     & \third{82.53±0.77}  & \second{74.03±0.87}  & 87.18±0.56  & \third{82.44±0.99}\\
               & LPE-T     & 78.45±0.98  & 73.40±1.14  & 87.79±0.63  & 57.10±0.60\\ \midrule
enron10 & No PEs   & 84.24±1.12  & \first{92.17±0.29}  & \first{94.34±0.29}  & 58.34±0.73\\
               & SLPE-E     & \first{90.76±0.37}  & 84.92±1.20  & \second{94.08±0.41}  & \second{91.52±1.12}\\
               & LPE-E     & 88.36±0.71  & \third{91.04±0.55}  & \third{93.98±0.30}  & 88.44±0.56\\
               & SLPE-I     & \third{89.75±0.83}  & \second{91.24±0.40}  & 93.96±0.55  & \first{93.19±0.43}\\
               & LPE-I     & 86.51±0.72  & 85.29±0.73  & 93.97±0.20  & \third{89.14±0.38}\\
               & SLPE-T     & \second{90.38±0.59} & 86.61±1.13  & 93.47±0.90  & 86.78±1.61\\
               & LPE-T     & 86.68±1.02  & 84.67±0.55  & 93.48±0.34  & 68.76±0.79\\ \bottomrule
\end{tabular}
\end{table}

\begin{table}[t]
\caption{AUC (\%) Average performance per variant across all datasets, models, and features.}\label{app:results-avg-per-variant}
\centering
\begin{tabular}{lc}
\toprule
Variant & Average Performance \\
\midrule
SLPE-I & \first{86.96} \\
SLPE-E & \second{86.38} \\
LPE-E & \third{86.21} \\
LPE-I & 84.72 \\
SLPE-T & 83.98 \\
LPE-T & 83.13 \\
No PEs & 78.13 \\
\bottomrule
\end{tabular}
\end{table}

\begin{table}[t]
\caption{AUC (\%) Average and median performance differences between E and I variants per model and between SLPE and LPE variants.}\label{app:results-diff-per-model}
\centering
\begin{tabular}{lcccc}
\toprule
Model & \multicolumn{2}{c}{E - I} & \multicolumn{2}{c}{SLPE - LPE} \\
\cmidrule(lr){2-3} \cmidrule(lr){4-5}
& Avg. Diff. & Median Diff. [Q1, Q3] & Avg. Diff. & Median Diff. [Q1, Q3] \\
\midrule
EGCN & 0.36 & 0.17 [-0.68, 0.57] & 3.91 & 2.27 [0.70, 4.86] \\
GRUGCN & 1.25 & 0.15 [-0.11, 1.43] & -0.30 & -0.01 [-1.16, 0.69] \\
HTGN & 0.003 & 0.02 [-0.08, 0.43] & 0.08 & 0.04 [-0.21, 0.23] \\
SLATE & 0.21 & 0.08 [-0.32, 0.36] & 0.66 & 0.46 [-0.02, 3.77] \\
\midrule
Mean Diff &0.46& 0.04&1.09&0.69 \\
\bottomrule
\end{tabular}
\end{table}

\begin{table*}
\caption{Average AUC (\%) performance comparison across datasets. The top two scores for each model and feature combination are highlighted by \space \first{First} and \space \second{Second}.}
\label{tab:avg-variant-feat-model}
\begin{minipage}[t]{0.48\textwidth}
\centering
\begin{tabular}{llccc}
\toprule
Model & Variant & one-hot & randn & zeros \\
\midrule
EGCN & No PEs & 88.01 & 81.43 & 50.0 \\
     & LPE-E   & 88.12 & 80.59 & 85.36 \\
     & LPE-I   & 87.87 & 73.74 & 85.2 \\
     & LPE-T   & 85.64 & 80.05 & 86.1 \\
     & SLPE-E   & \first{88.93} & 83.27 & \second{88.79} \\
     & SLPE-I   & \second{88.88} & \first{87.62} & \first{89.55} \\
     & SLPE-T   & 87.78 & \second{85.08} & 87.92 \\
\midrule
GRUGCN & No PEs & 84.78 & 79.68 & 84.14 \\
       & LPE-E   & \first{88.02} & \first{83.68} & \first{88.96} \\
       & LPE-I   & 87.04 & 80.4  & 84.56 \\
       & LPE-T   & 85.88 & 79.14 & 80.83 \\
       & SLPE-E   & \second{87.2} & 80.64 & 85.4 \\
       & SLPE-I   & 87.08 & \second{81.17} & \second{86.16} \\
       & SLPE-T   & 85.3  & 80.31 & 82.54 \\
\bottomrule
\end{tabular}
\end{minipage}
\hfill
\begin{minipage}[t]{0.48\textwidth}
\centering
\begin{tabular}{llccc}
\toprule
Model & Variant & one-hot & randn & zeros \\
\midrule
HTGN & No PEs & 92.44 & \first{91.08} & 85.37 \\
     & LPE-E   & \second{92.71} & 89.39 & 89.55 \\
     & LPE-I   & 92.53 & 89.2  & 88.99 \\
     & LPE-T   & 89.34 & 87.52 & \first{89.82} \\
     & SLPE-E   & \first{92.73} & 88.94 & 89.28 \\
     & SLPE-I   & 92.57 & \second{89.52} & \second{89.76} \\
     & SLPE-T   & 90.52 & 86.99 & 89.44 \\
\midrule
SLATE & No PEs & 85.29 & 59.81 & 55.49 \\
      & LPE-E   & \first{86.14} & 80.2  & 81.83 \\
      & LPE-I   & 85.41 & 80.49 & 81.23 \\
      & LPE-T   & 84.94 & 68.44 & 79.87 \\
      & SLPE-E   & \second{86.1} & \second{81.44} & \first{83.86} \\
      & SLPE-I   & 85.53 & \first{82.25} & \second{83.43} \\
      & SLPE-T   & 85.86 & 72.96 & 73.05 \\
\bottomrule
\end{tabular}
\end{minipage}
\end{table*}

\begin{table}[t]
\caption{AUC (\%) Average and Median difference with IQR for the difference between 'Best PE' and 'No PEs'.}
\label{tab:performance_diff}
\centering
\begin{tabular}{lccc}
\hline
 Model &Avg. Diff. & Median Difference (IQR) \\
\hline
EGCN   &15.95 & $8.20 \,  [1.50, 36.06]$  \\
GRUGCN &4.23 & $2.35 \,   [0.87, 6.19 ]$   \\
HTGN   &1.40 & $-0.29 \,   [-0.47, 0.75]$     \\
SLATE  &18.05 & $16.63 \, [0.22, 34.23]$ \\
\hline
\end{tabular}
\end{table}

\begin{table}[t]
\caption{AUC (\%) Average and median performance differences between Best PE and No PEs by feature for each model, with mean and standard deviation, and median with quartiles.}
\label{tab:diff-by-feature-model}
\centering
\begin{tabular}{llcccc}
\toprule
Feature & Model & \multicolumn{2}{c}{Mean $\pm$ Std} & \multicolumn{2}{c}{Median Diff. [Q1, Q3]} \\
\midrule
one-hot
  & EGCN   & 1.50 & 1.37 & 1.48  & [1.03, 1.95] \\
  & GRUGCN & 3.34 & 3.02 & 2.21  & [1.77, 3.79] \\
  & HTGN   & 0.34 & 1.18 & -0.08 & [-0.48, 0.75] \\
  & SLATE  & 0.98 & 1.85 & 0.12  & [-0.02, 1.13] \\
\midrule
random
  & EGCN   & 6.70  & 5.77  & 8.20  & [4.56, 10.35] \\
  & GRUGCN & 4.39  & 4.93  & 3.95  & [1.46, 6.88]  \\
  & HTGN   & -1.29 & 1.98  & -0.35 & [-1.39, -0.25] \\
  & SLATE  & 23.66 & 16.06 & 29.83 & [20.06, 33.42] \\
\midrule
constant
  & EGCN   & 39.64 & 4.13  & 38.97 & [36.79, 41.83] \\
  & GRUGCN & 4.96  & 7.58  & 2.02  & [-0.03, 7.01]  \\
  & HTGN   & 5.13  & 6.85  & 3.65  & [-0.35, 9.13]  \\
  & SLATE  & 29.49 & 15.45 & 36.32 & [27.16, 38.65] \\
\bottomrule
\end{tabular}
\end{table}

\end{document}